\documentclass[conference]{IEEEtran}
\IEEEoverridecommandlockouts

\usepackage{cite}
\usepackage{amsmath,amssymb,amsfonts}
\usepackage{graphicx}
\usepackage{textcomp}
\usepackage{xcolor}
\usepackage{booktabs}
\usepackage{multirow}
\usepackage{array}
\usepackage{tabularx}
\usepackage{float}
\usepackage{url}
\usepackage[british]{babel}
\usepackage{csquotes}
\usepackage{hyperref}

\title{Per-Aetiology Contrastive Severity Embeddings with Phonological
Pseudo-Labelling for Multilingual Dysarthric Speech}

\author{%
\IEEEauthorblockN{Bernard Muller\IEEEauthorrefmark{1}\thanks{Accepted at the 2026 IEEE Spoken Language Technology Workshop (SLT 2026). \copyright~2026 IEEE. Personal use of this material is permitted. Permission from IEEE must be obtained for all other uses, in any current or future media, including reprinting/republishing this material for advertising or promotional purposes, creating new collective works, for resale or redistribution to servers or lists, or reuse of any copyrighted component of this work in other works.},
Antonio Armando Ortiz Barra\~n\'on\IEEEauthorrefmark{2},
LaVonne Roberts\IEEEauthorrefmark{1}\IEEEauthorrefmark{3}}%
\IEEEauthorblockA{\IEEEauthorrefmark{1}The Scott-Morgan Foundation, Torquay, United Kingdom}%
\IEEEauthorblockA{\IEEEauthorrefmark{2}Tecnol\'ogico de Monterrey, Monterrey, Mexico}%
\IEEEauthorblockA{\IEEEauthorrefmark{3}SMF Labs, Paris, France --- Corresponding author: \texttt{lavonne@scottmorganfoundation.org}}%
}

\begin{document}
\maketitle

\begin{abstract}
Most multilingual dysarthria-severity systems either train on a single aetiology-language pair or pool heterogeneous aetiologies into one label space. We test that pooling assumption with four matched HuBERT-base contrastive embedding models under a shared backbone, training recipe, corpus registry and held-out evaluation: one mixed-aetiology baseline and three aetiology-specific models for cerebral palsy (CP), Parkinson's disease (PD) and amyotrophic lateral sclerosis (ALS). Training combines clinically labelled speech with ordinal pseudo-labels from a training-free phonological profiling method [1], [2]. On speaker-disjoint, leakage-filtered held-out subsets, the per-aetiology models outperform the mixed baseline across all three target aetiologies: CP (macro F1 0.829 vs 0.676, +22.6 \% relative), PD (0.715 vs 0.511, +40.0 \%) and ALS (0.788 vs 0.596, +32.3 \%). On CP, adding 144 SAP and 44 CDSD pseudo-labelled speakers lifts macro F1 from 0.786 to 0.829 over a clinical-only CP model (+4.3 percentage points). Training data span three to seven languages per aetiology. We position this as a controlled comparison of label-space design choices and discuss pseudo-label calibration, split hygiene, and confidence-thresholded deployment as important limitations for future work.
\end{abstract}

\section{Introduction}
Dysarthria --- a group of motor speech disorders arising from neurological damage to the speech production mechanism --- has been reported at roughly 170 per 100,000 people in the United Kingdom [3] and is a leading symptom of Parkinson's disease (PD), amyotrophic lateral sclerosis (ALS), cerebral palsy (CP) and several other aetiologies. Automated severity estimation supports longitudinal monitoring of disease progression, conditioning of dysarthric automatic speech recognition (ASR) and adaptation of augmentative and alternative communication (AAC) devices. Published systems fall into two broad groups: single-aetiology, single-language classifiers trained on a specific corpus (typically UA-Speech for CP or a national PD corpus), and larger mixed-aetiology models trained on whatever labelled data is available, most recently Google's SpICE [4] on 551 k English utterances. The implicit assumption in the latter is that severity is a single learnable construct and that more data --- regardless of aetiology --- will improve it. We test that assumption directly and show that, under matched architecture and training, aetiology-specific severity training beats pooled training across CP, PD, and ALS, and that training-free phonological pseudo-labels strengthen classification within an aetiology.

We test this pooling assumption empirically under a controlled architecture, training, and evaluation setup. Clinical severity labels are heterogeneous across aetiologies and corpora, and their numerical meanings are not directly interchangeable: CP intelligibility percentage [5], the Unified Parkinson's Disease Rating Scale, and the ALS Functional Rating Scale measure different constructs and are not directly comparable. Phonological degradation patterns differ by aetiology too: PD produces hypokinetic dysarthria with minimal loss of consonant contrast until late stages, whereas CP produces spastic dysarthria with early collapse of stridency and manner contrasts [1], [2]. A single supervised model forced to reconcile these scales and patterns cannot do so consistently.

{\sloppy\hbadness=10000 Our paper isolates one design choice: aetiology-disaggregated training. We report its effect on severity-classification performance while holding everything else fixed: architecture, training procedure, corpus registry, and evaluation. We train four contrastive embedding models on the same HuBERT-base backbone, using the same three-stage recipe (binary healthy-dysarthric, ordinal severity with hard-negative mining, cross-lingual alignment): one each for CP, PD, and ALS, plus one trained on all aetiologies jointly. Training data are drawn from a multilingual dysarthric-corpus registry building on [2], augmented with large healthy-speech baselines for class balance and extended with training-free phonological pseudo-labels [1] on the minority severity classes (expansion factors per class and aetiology are given in the supplementary material). All four models are then evaluated on speaker-disjoint, leakage-filtered Phase 2 held-out test pools restricted to the target aetiology plus language-matched healthy controls. Per-aetiology and mixed-baseline pools differ slightly in size because they were drawn from each model's own training, validation, and test partition; both share the same registry, the same speaker-disjoint protocol, and the same per-aetiology and HC-matching filters, so the comparison remains within a single sampling design.\par}

Our contributions are:

Evidence that per-aetiology training dominates the mixed baseline across all three aetiologies tested --- CP +22.6 \% relative macro F1 (0.829 vs 0.676), PD +40.0 \% (0.715 vs 0.511), ALS +32.3 \% (0.788 vs 0.596) --- on clean, speaker-disjoint, leakage-filtered test subsets.

To our knowledge, this is among the first controlled cross-lingual comparisons of aetiology-specific severity classifiers for PD and ALS under a matched embedding architecture. PD training covers seven languages (Slovak, Dutch, Spanish, Italian, Hungarian, Portuguese, English) and ALS five (English, Italian, German, Portuguese, French). Recent multilingual ALS severity work also exists [15]; the most closely-matching prior systems are otherwise single-corpus and monolingual. We note that the held-out test set is heavily English- and SAP-concentrated (\S{}4.6), so the cross-lingual claim applies primarily to training coverage; balanced multilingual evaluation is a separate step we discuss under limitations.

Phonological pseudo-labelling from a training-free source [2] lifts CP macro F1 by +4.3 points over a clinical-only CP model. The same pseudo-label source expands PD and ALS severity coverage, including settings and languages for which cross-lingual dysarthric-severity evidence remains sparse.

The rest of the paper reviews related work (\S{}2), describes the data and training recipe (\S{}3), reports the per-aetiology comparison and robustness checks (\S{}4), and closes with discussion and conclusions (\S{}5--6).

\section{Related Work}
\textbf{Supervised severity classification.} Kadirvelu et al. [6] introduced SALR --- a wav2vec2 + triplet-loss model trained on UA-Speech --- reporting 70.48 \% leave-one-speaker-out (LOSO) accuracy (macro F1 0.593) on CP. The evaluation is single-corpus and English-only. Yeo et al. [7] trained an XGBoost classifier on 39 handcrafted features spanning voice quality, pronunciation and prosody, achieving 0.671 macro F1 across English, Korean and Tamil; the approach is cross-lingual but relies on handcrafted descriptors rather than learned speech representations. Venugopalan et al. [4] trained SpICE on 551 k English utterances from Project Euphonia. A preliminary cross-lingual evaluation of the released SpICE model on a five-language subset of our registry showed substantially reduced healthy-speaker recall outside English (details in the supplementary material), consistent with the expectation that English-only training does not transfer cleanly across languages.

\textbf{Pseudo-labelling and contrastive learning for severity.} Bae et al. [8] combine pseudo-labelled dysarthric samples with SSL contrastive pre-training and report Spearman correlation 0.761 on severity regression --- strong evidence that pseudo-labels add value --- but the work is monolingual and addresses regression, not ordinal classification. Our approach differs in two ways: (i) we source pseudo-labels from a training-free phonological method [1] that requires only healthy speech plus forced alignment, so it scales cross-lingually without any labelled dysarthric audio; and (ii) we train separate models per aetiology rather than one shared model.

\textbf{Phonological subspace severity assessment.} Muller et al. [2] showed that d-prime scores along nine consonant and vowel phonological feature directions in frozen HuBERT embeddings [9] produce a severity-correlated, aetiology-specific and cross-lingually stable profile across 3,374 speakers, 25 corpora and 12 languages, extending the earlier 890-speaker / 10-corpus / 5-language training-free severity assessment in [1]. Aetiology profiles are preserved across languages (cosine > 0.95) and aetiology-specific profiles emerge without supervision, which is why the method is a natural pseudo-label source for under-resourced languages. We use it here to pseudo-label 1,181 previously unlabelled dysarthric speakers across five corpora and five languages: 1,041 SAP [16] English speakers (419 PD, 239 ALS, 144 CP, 144 Down syndrome, 95 stroke), 44 CDSD [20] Mandarin CP speakers, 71 EWA-DB [19] Slovak PD speakers, 18 Hungarian\_Dysarthria (HUNDYSDB [31]) mixed-aetiology speakers and 7 AVFAD [24] Portuguese speakers (4 ALS, 3 PD). To our knowledge, this is the first use of the training-free d-prime source to provide useful ordinal pseudo-labels for ALS and Slavic-family PD at this scale.

\textbf{Aetiology pooling in dysarthric speech research.} Most recent large-scale dysarthric models either restrict to a single aetiology-language pair [6], [7] or pool aetiologies implicitly (e.g., SpICE [4] mixes CP, PD, ALS, stroke and other aetiologies under a single ``dysarthric'' label). To our knowledge, no prior paper has systematically compared per-aetiology versus mixed-aetiology training with controlled architecture, training recipe and evaluation --- which is the gap we address.

\textbf{Self-supervised speech representations.} HuBERT [9] and wav2vec 2.0 [10] provide the representational backbone for most recent severity systems. We use HuBERT-base with the top four transformer layers unfrozen --- sufficient to adapt to dysarthric acoustics while keeping the 95 M-parameter model tractable for academic GPUs.

\section{Method}
\subsection{Data}
Stage 1 is trained on 1,429,582 samples from 30 source corpora across 10 languages, comprising 19 dysarthric or mixed corpora plus 11 large healthy-speech baselines that provide class-balanced negatives for the binary healthy-vs-dysarthric objective. Stages 2--3 restrict training to ten dysarthric corpora with clinical severity labels --- TORGO [17], UA-Speech [18], COPAS, SSNCE-Tamil [21], VOC-ALS [25], MDSC [26], IPVS, PC-GITA [30], Neurovoz [23] and the SAP clinical subset --- plus the pseudo-labelled subsets in Table II. Portions of the PD subsets used in this study were obtained through the OneVoice-MSD 2026 initiative [29] and are used under its research-only Data-Use Agreement; per that agreement, we cite Hernandez et al.~[29] and acknowledge the SLT-2026 OneVoice-MSD Focused Track. Table I summarises the four aetiology-specific training configurations evaluated in this paper.

\begin{table}[t]\centering\scriptsize
\caption{Training configurations. All four models use clinical + training-free phonological pseudo-labels (Sec.~3.4). Dysarthric-speaker counts are shown; total training speakers additionally include healthy controls drawn from matched corpora.}
\setlength{\tabcolsep}{3pt}
\resizebox{\columnwidth}{!}{%
\begin{tabular}{lrrrrrr}
\toprule
\textbf{Model} & \textbf{Train} & \textbf{Val} & \textbf{Test} & \textbf{Dys. spk.} & \textbf{Langs (train)} & \textbf{Aetiology} \\
\midrule
CP (ours) & 313,820 & 39,228 & 39,227 & 406 & 3: en, ta, zh & CP + HC \\
PD (ours) & 189,793 & 23,724 & 23,724 & 638 & 7: en, es, nl, sk, it, hu, pt & PD + HC \\
ALS (ours) & 129,495 & 16,187 & 16,187 & 403 & 5: de, en, fr, it, pt & ALS + HC \\
Mixed (baseline) & 727,754 & 90,969 & 90,969 & 1,831 & 11 & all mixed \\
\bottomrule
\end{tabular}
}
\end{table}

\textbf{Healthy-control matching.} Healthy controls are language-matched, capped to the size of the largest dysarthric severity class per configuration, and stratified-downsampled across languages in proportion to dysarthric sample counts. Controls from languages without any target-aetiology dysarthric speaker are excluded. This caps HC at 28--42 \% of each training set, prevents HC from dominating the contrastive loss, and avoids a language-as-severity shortcut. We follow the per-language HC strategy used in [1].

\textbf{Shared test protocol.} All four models are evaluated under the same registry and filtering protocol drawn from the Phase 2 filtered test set of [2] --- recordings in languages that contain at least one dysarthric speaker, so that healthy-control samples do not inflate accuracy in cross-lingual models. The Phase 2 test set comprises 89,068 samples across eight languages with dysarthric data (en, zh, ta, nl, hu, es, it, sk) after applying speaker-disjoint and leakage filters (see \S{}3.5). For the per-aetiology comparison we further restrict to speakers of the target aetiology plus healthy controls in matched languages. The mixed and per-aetiology models are scored on slightly different filtered test pools derived from each model's own train/validation/test partition; both share the same registry, the same speaker-disjoint protocol and the same per-aetiology and HC-matching filters, so the comparison remains within a single sampling design.

\textbf{Severity harmonisation.} Severity harmonisation is necessarily approximate. Where source corpora provided intelligibility percentages (UA-Speech, TORGO, SAP), we mapped them to four levels using the Stipancic et al. [5] thresholds: $\geq$ 94 \% = control, 85--94 \% = mild, 70--84 \% = moderate, < 70 \% = severe. Where source corpora already provided four-level severity categories, we retained those mappings directly. Cross-aetiology comparability of the resulting labels remains imperfect, as discussed in [2] \S{}5.2; the labels support quantitative comparison within this study but should not be interpreted as a perfectly calibrated cross-corpus clinical severity scale.

\textbf{Pseudo-labelled subsets.} Five corpora contribute pseudo-labelled speakers via the training-free phonological method of [1], [2], covering all three target aetiologies:

\begin{table}[t]\centering\scriptsize
\caption{Pseudo-labelled subsets across five corpora and five languages; speakers labelled via the training-free phonological method of [1], [2] using 11 of the 15 features defined in [2] (see Sec.~3.4 for the exclusion criteria).}
\setlength{\tabcolsep}{3pt}
\resizebox{\columnwidth}{!}{%
\begin{tabular}{lrrr}
\toprule
\textbf{Corpus} & \textbf{Language} & \textbf{Speakers} & \textbf{Aetiology breakdown} \\
\midrule
SAP & English & 1,041 & PD 419, ALS 239, CP 144, DS 144, stroke 95 \\
CDSD & Mandarin & 44 & CP 44 \\
EWA-DB & Slovak & 71 & PD 71 \\
Hungarian\_Dysarthria & Hungarian & 18 & mixed (PD/stroke) \\
AVFAD & Portuguese & 7 & ALS 4, PD 3 \\
Total &  & 1,181 &  \\
\bottomrule
\end{tabular}
}
\end{table}

Internal comparison against clinically labelled subgroups within the same corpora indicates close agreement (cosine $>$ 0.95) between pseudo-label and clinical-label group-mean aetiology profiles, suggesting that pseudo-labels reproduce the same phonological degradation signatures as expert annotations at the group level.

\subsection{Architecture}
The embedding network comprises a HuBERT-base backbone [9], [27] (95 M parameters, 12 transformer layers, 768-dim hidden state) with the top four transformer layers unfrozen. Frame-level hidden states from the final layer are mean-pooled to produce a 768-dim utterance vector, which is passed through a two-layer MLP projection head (768 $\rightarrow$ 512 $\rightarrow$ 256) with GELU activation and dropout 0.1. The output is L2-normalised to a 256-dim embedding on the unit hypersphere.

Downstream classification uses a linear probe trained on the frozen embeddings. Probe choice turned out to matter: early evaluations used class\_weight="balanced", which re-weights rare classes to match the frequency of the dominant class. On imbalanced training distributions (e.g., ALS with 1,501 mild / 207 moderate / 231 severe / 151 control samples), this interacted pathologically with non-uniform between-class separability and collapsed the moderate class (see supplementary material). We therefore default to an unweighted linear probe (sklearn LogisticRegression [28], max\_iter=1000, random\_state=42). A k-nearest-neighbour probe (k = 15) is reported as an unbiased sanity check --- kNN does not apply class weighting and produces a readout of embedding-space separability that is independent of the probe-weighting question.

\subsection{Three-stage contrastive training}
All models share the same Stage 1 initialisation and differ only in the aetiology-specific data used in Stages 2a--3.

\textbf{Stage 1 --- Binary healthy vs dysarthric (InfoNCE)} [11]. A single backbone is trained on the full training mix across all aetiologies to separate healthy from dysarthric speech. The resulting stage1\_best.pt is reused as initialisation for all four models. Batch size 256, 5 epochs, AdamW 1e-4.

\textbf{Stage 2a --- Pairwise boundary contrastive.} Positives are same-severity, same-language pairs. Hard negatives are severity-boundary pairs (mild--moderate, moderate--severe). InfoNCE temperature 0.07; hard negatives are weighted 3$\times$ in the denominator. This stage is the largest single contributor to macro-F1 improvement (see \S{}4.5).

\textbf{Stage 2b --- Ordinal severity triplet} [12]. Anchor--positive pairs are drawn from the same severity class; negatives from adjacent classes receive a small margin (0.2) and from distant classes a larger margin (0.4), implementing an ordinal penalty proportional to rank distance.

\textbf{Stage 3 --- Cross-lingual contrastive alignment.} Batches are balanced across languages within each severity class. A cross-lingual InfoNCE loss encourages same-severity samples from different languages to align; the ordinal triplet loss from Stage 2b is retained as a regulariser.

Each stage trains for up to 5 epochs with early stopping on validation macro F1 (patience 2). Total training time for all four models was \textasciitilde{}68 h on a single H100.

\subsection{Phonological pseudo-labelling}
\begin{figure}[t]\centering
\includegraphics[width=\linewidth]{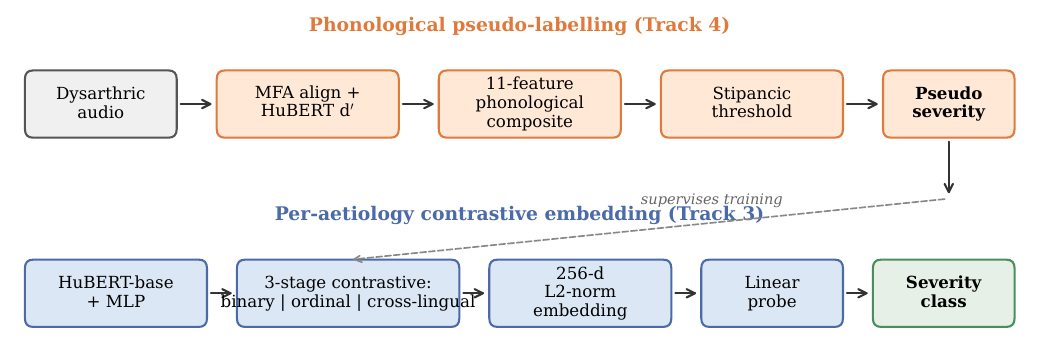}
\caption{Two-stage pipeline: training-free phonological pseudo-labelling (Stage 1) labels unlabelled speakers; per-aetiology contrastive HuBERT embeddings (Stage 2) are then trained on the expanded corpus.}
\label{fig:1}
\end{figure}

We briefly summarise the pseudo-labelling procedure; full technical detail is in [1], [2]. For each speaker: (i) MFA [13] phone alignment produces time-stamped IPA intervals; (ii) HuBERT-base embeddings are mean-pooled per phone; (iii) nine phonological feature directions (5 consonant, 4 vowel) are computed from healthy controls of the same language; (iv) d-prime scores along each direction form an 11-feature profile (9 d-prime + boundary sharpness + cross-position cosine); (v) a composite score derived from a regression calibrated on a held-out labelled subset is thresholded into the four Stipancic severity bins. This 11-feature profile is a subset of the 15-feature phonological/prosodic profile defined in [2] \S{}3.1; we exclude the vowel triangle area (VTA), which requires three tokens each of /a/, /i/, /u/ per speaker and is available for only $\sim$ 10\,\% of the SAP candidate pool, and the three prosodic features, which were not in the calibrated regression. The 11-feature subset keeps the calibration set large enough to estimate the moderate/severe threshold. The Stipancic mapping is consistent with the procedure in [1], applied here within rather than across corpora. Internal comparison against clinical labels in shared SAP, MDSC and EWA-DB subsets indicates close agreement (cosine $>$ 0.95) between pseudo-label and clinical-label group-mean aetiology profiles; this is consistent with, rather than identical to, the cross-lingual profile-shape stability reported in [2] (supplementary Table~X). As a stricter independent check, on the OneVoice-MSD 2026 Czech, German and Spanish PD subsets (415 speakers, clinical labels independent from the pseudo-label pipeline), linearly-weighted Cohen's $\kappa$ between pseudo-labels and clinical labels was 0.021 / 0.036 / 0.178 (pooled $\kappa$ 0.069). The pseudo-labels are therefore a training-time supervisory signal, not a clinical estimate: their utility for training (\S{}4.2) does not require utterance-level agreement with a specific clinical rubric, since the quantile-mapped composite score is calibrated at the corpus level rather than to any external ordinal scale. Cross-lingual transfer of the thresholds themselves is not claimed: pseudo-labels are treated strictly as within-corpus ordinal supervision. Concretely, [1] and [2] caution that absolute d-prime magnitudes are not cross-lingually calibrated, so we use the composite score's quantile structure (not its absolute value) to map to Stipancic severity bins, and restrict pseudo-labels to corpora in which calibration was validated against a within-corpus clinical reference (SAP, CDSD, EWA-DB, Hungarian\_Dysarthria, AVFAD).

\begin{figure}[t]\centering
\includegraphics[width=\linewidth]{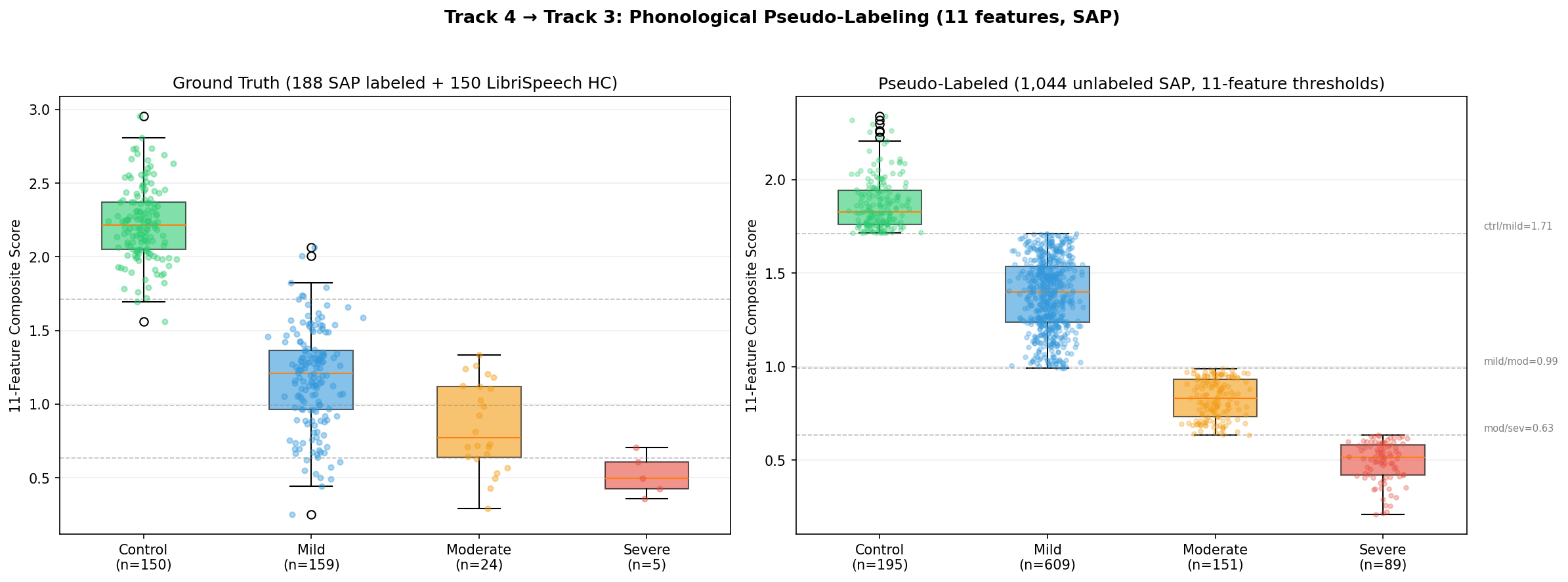}
\caption{Pseudo-label threshold calibration on SAP. Left: 11-feature composite d-prime score by clinical severity on a labelled validation subset (188 SAP + 150 LibriSpeech HC), used to derive the control/mild, mild/moderate, and moderate/severe thresholds (1.71, 0.99, 0.63). Right: resulting pseudo-label distribution on 1,044 previously unlabelled SAP speakers (1,041 of which are dysarthric across five aetiologies as listed in Sec.~3.4; the remaining three pseudo-label to control). The labelled distribution reproduces the ordinal severity structure expected under the Stipancic thresholds.}
\label{fig:2}
\end{figure}

\subsection{Speaker-disjoint enforcement}
All evaluation numbers are reported after enforcing speaker-disjoint splits across every source corpus. Where the shipped manifest is not speaker-disjoint --- notably UA-Speech, which distributes 15 dysarthric and 13 control speakers across train and test at the utterance level --- we remove overlapping speakers from both probe splits. Single-split corpora are held out as zero-shot test sources and are not used for the headline comparison.

\section{Results}
\subsection{Per-aetiology vs mixed-aetiology}
Table III reports the headline result. On each aetiology's held-out test subset (the target aetiology plus healthy controls in matched languages, speaker-disjoint and leakage-filtered), the per-aetiology model beats the mixed baseline by a large margin across all three aetiologies.

\begin{table}[t]\centering\scriptsize
\caption{Per-aetiology vs mixed-aetiology on each aetiology's held-out test subset. All probes are unweighted linear (class\_weight=None).}
\setlength{\tabcolsep}{3pt}
\resizebox{\columnwidth}{!}{%
\begin{tabular}{lrrrrrr}
\toprule
\textbf{Aetiology} & \textbf{Model} & \textbf{N-test} & \textbf{Acc} & \textbf{Macro F1} & \textbf{$\Delta$ F1} & \textbf{Rel.} \\
\midrule
CP & Mixed & 52,652 & 73.6 \% & 0.676 & --- & --- \\
CP & CP (ours) & 49,755 & 88.5 \% & 0.829 & +0.153 & +22.6 \% \\
PD & Mixed & 29,242 & 84.8 \% & 0.511 & --- & --- \\
PD & PD (ours) & 36,325 & 93.2 \% & 0.715 & +0.204 & +40.0 \% \\
ALS & Mixed & 25,217 & 89.5 \% & 0.596 & --- & --- \\
ALS & ALS (ours) & 25,217 & 95.0 \% & 0.788 & +0.192 & +32.3 \% \\
\bottomrule
\end{tabular}
}
\end{table}

The gain is not attributable to a single class. Per-class F1 (Table IV) shows that the per-aetiology models improve on mild, moderate and severe simultaneously; the control class is already well-separated by all models because Stage 1 trains a dedicated healthy-vs-dysarthric boundary. The PD result is the largest in relative terms --- Mixed's mixed-aetiology severity space cannot support a reliable mild-versus-moderate decision for hypokinetic dysarthria, where the degradation signature is weak; PD, trained only on PD speakers, achieves a cleaner boundary.

\subsection{Pseudo-labels add value within an aetiology, not across}
We can test the contribution of pseudo-labels only for CP. Clinical CP labels alone total \textasciitilde{}23 k samples (mainly UA-Speech, TORGO, MDSC and SSNCE\_Tamil), which are enough to train a CP-only model for comparison: the clinical-only CP model reaches 0.786 macro F1 on its own held-out CP test. Adding 144 SAP CP pseudo-labelled speakers plus 44 CDSD Mandarin CP pseudo-labelled speakers lifts macro F1 to 0.829 --- a +4.3 percentage-point gain (+5.5 \% relative) from pseudo-labelled data alone, without any additional clinical annotation.

For PD and ALS the equivalent clinical-only baseline cannot be trained: the clinical-labelled PD and ALS data in our registry (primarily EasyCall [22] PD, VOC-ALS and scattered clinically-annotated SAP subsets) are too sparse to support a meaningful per-aetiology model, which is precisely why pseudo-labels are needed for those aetiologies in the first place. The PD and ALS per-aetiology models in Table I are therefore trained on clinical + pseudo combined, with pseudo-labels providing the bulk of moderate-class and cross-lingual coverage. Their +40.0 \% and +32.3 \% relative gains over the mixed baseline show that the combined training produces better aetiology-specific severity representations than mixing aetiologies, but they do not isolate the pseudo-label contribution from the per-aetiology contribution.

\subsection{Per-class analysis and the probe-weighting question}
Table IV breaks down macro F1 into its four severity components for each aetiology. The per-aetiology model improves every class (modulo a small drop on control that is already near ceiling).

\begin{table}[t]\centering\scriptsize
\caption{Per-class F1 on each aetiology's held-out test subset.}
\setlength{\tabcolsep}{3pt}
\resizebox{\columnwidth}{!}{%
\begin{tabular}{lrrrrrr}
\toprule
\textbf{Aetiology} & \textbf{Model} & \textbf{Control} & \textbf{Mild} & \textbf{Moderate} & \textbf{Severe} & \textbf{Macro} \\
\midrule
CP & Mixed & 0.93 & 0.75 & 0.57 & 0.39 & 0.676 \\
CP & CP & 0.95 & 0.86 & 0.69 & 0.82 & 0.829 \\
PD & Mixed & 0.94 & 0.70 & 0.31 & 0.22 & 0.511 \\
PD & PD & 0.97 & 0.80 & 0.55 & 0.54 & 0.715 \\
ALS & Mixed & 0.97 & 0.82 & 0.16 & 0.44 & 0.596 \\
ALS & ALS & 0.98 & 0.84 & 0.59 & 0.64 & 0.788 \\
\bottomrule
\end{tabular}
}
\end{table}

{\sloppy\hbadness=10000 A separate probe-choice experiment (reported in the supplementary material) illustrates why an unweighted probe is the right default on imbalanced training. On ALS, class\_weight="balanced" upweights the 207 moderate and 231 severe training samples by about $7\times$ to match the 1,501 mild samples. Given the modest between-class separability of ALS moderate versus severe (centroid cosine 0.44), the probe collapses moderate into severe (moderate F1 falls to 0.038) and over-predicts severe (recall 1.00, precision 0.32). Removing the class weighting recovers moderate F1 to 0.589 and improves overall macro F1 from 0.586 to 0.788. Unweighted linear and kNN-15 give essentially the same result, confirming that the moderate class is present and separable in the embedding space --- the balanced probe simply misclassified it.\par}

\subsection{Embedding structure visualisation}
Per-class silhouette and t-SNE projections (supplementary material) show coherent ordinal severity progressions in each per-aetiology embedding space, with substantial band overlap between mild and moderate for the mixed baseline, consistent with its lower macro F1.

\subsection{Robustness and efficiency analyses}
These analyses hold out speakers or corpora at the linear-probe level; the HuBERT backbone saw the full registry, so results are indicative rather than a strict unseen-speaker / unseen-corpus test. \textbf{LOSO on UA-Speech} matching the SALR setup [6] gives 80.0\,\% accuracy / 0.539 macro F1 vs SALR's 70.48\,\% / 0.593 --- +9.5\,pp absolute on accuracy; the lower macro F1 reflects three misclassifications in two-active-class folds that penalise the metric disproportionately. \textbf{LOCO on CP} (retrained linear probe on N$-$1 corpora, tested on the held-out corpus): per-corpus macro F1 UA-Speech 0.947, SSNCE\_Tamil 0.518, SAP 0.444, MDSC 0.436, CDSD 0.346, TORGO 0.343 (mean 0.506 $\pm$ 0.226, sample SD). The UA-Speech--SAP gap of 0.503 exceeds the per-aetiology CP-vs-mixed gap, indicating that recording style dominates cross-corpus error over language. \textbf{Stage ablation and confidence-thresholded coverage} (validation-set stage-wise improvement; softmax $\geq$ 0.9 retains 64\,\% of test samples at 95.2\,\% accuracy and 0.924 macro F1) are reported in full in the supplementary material.

\subsection{Common-subset (matched-data) confirmation}
Because each model is scored on its own filtered Phase~2 pool, pool sizes differ mechanically with the language-matched-HC filter (\S{}3.1). To rule out that the Table~III gap is an artefact of pool composition, we re-score both models on the strict speaker-level intersection of the two pools (Table~V). Direction survives in 5 of 6 (probe~$\times$~aetiology) cells and never reverses; the PD linear tie ($\Delta$ +0.007) reflects zero moderate--severe speakers in the PD common subset, an artefact both models absorb, and the kNN-15 readout on the same subset recovers $\Delta$ +0.067.

\begin{table}[t]\centering\scriptsize
\caption{Common-subset (matched-data) re-scoring on the speaker-level intersection of per-aetiology and mixed test pools. Same probe as Table~III.}
\setlength{\tabcolsep}{3pt}
\resizebox{\columnwidth}{!}{%
\begin{tabular}{lrrrrr}
\toprule
\textbf{Aetiology} & \textbf{Common spk} & \textbf{Mixed F1} & \textbf{Per-aet F1} & \textbf{$\Delta$ linear} & \textbf{$\Delta$ kNN-15} \\
\midrule
CP  &  66 & 0.751 & 0.859 & +0.109 & +0.161 \\
PD  &  65 & --    & --    & +0.007 & +0.067 \\
ALS & 253 & 0.596 & 0.686 & +0.091 & --     \\
\bottomrule
\end{tabular}
}
\end{table}

\subsection{Mixed encoder with per-aetiology head}
To test whether the gain in Table~III is driven by the aetiology-specific representation rather than by having a separate classifier head, we train a per-aetiology linear head on top of the mixed encoder using each aetiology's held-out training partition, and evaluate on the target-aetiology test subset. Macro-F1: CP 0.692, PD 0.506, ALS 0.590. This is within 0.01 macro-F1 of the mixed-encoder + all-aetiologies linear-probe baseline in Table~III and 0.09--0.20 below the per-aetiology encoder in the same table; the same picture holds under a kNN-15 readout. The gain therefore lives in the encoder, not the head: pooling aetiologies degrades the representation itself, and swapping the head does not recover it.

\section{Discussion}
\textbf{Aetiology-specific training is the stronger default.} The three-way result --- CP +22.6\,\%, PD +40.0\,\%, ALS +32.3\,\% relative macro-F1 --- indicates that under a matched architecture and recipe, dysarthria severity is not a single learnable construct across aetiologies. The common-subset re-scoring (\S{}4.6) and the separate-heads ablation (\S{}4.7) together show this is not an artefact of test-pool composition or classifier-head design: the pooling penalty lives in the representation itself. Independent probe-level transfer to the SAND cohort (339 Italian ALS/HC speakers, supplementary material) further shows the ALS encoder beats the CP encoder by 0.07 macro-F1 out-of-corpus and out-of-language. This aligns with the Duffy [14] clinical taxonomy and the phonological evidence in [2] that PD-moderate and CP-moderate speakers produce opposite d-prime patterns. An operational pipeline must route patients to an aetiology-specific severity head by clinical diagnosis or via a preceding aetiology classifier such as [2]. The +4.3 pp CP gain from adding 144 SAP and 44 CDSD pseudo-labelled speakers (0.786 $\rightarrow$ 0.829) shows the training-free d-prime method [1] provides useful ordinal pseudo-labels within an aetiology; scaling to new languages requires matched healthy speech and within-corpus calibration.

\textbf{PD is the harder target.} PD macro F1 is 0.715 vs CP 0.829. Hypokinetic dysarthria preserves phonological contrasts until late stages --- PD severe speakers retain $\sim$90\,\% of healthy d-prime on voicing and sonorance [2] --- so the mild--control boundary is intrinsically ambiguous. Feature engineering beyond d-prime (prosodic rhythm, voice quality) may be required to close the PD--CP gap; nonetheless the +40\,\% relative improvement over the mixed baseline shows aetiology-specific training remains strictly better in this unfavourable regime.

\textbf{Limitations.} (i) Single training run per model; multi-seed replication was not completed and remains future work. Speaker-bootstrap CIs in the supplementary material characterise test-side uncertainty but not training-side seed variance. (ii) SAP contributes the majority of the ALS and PD held-out test set (supplementary \S{}S.11: 94\,\% and 67\,\%); cross-lingual results for these aetiologies reflect primarily English-language SAP material. A test-side SAP-excluded probe-level rerun (supplementary material) shows residual non-SAP subsets are too small to resolve a per-aetiology-vs-mixed comparison at Table~III's resolution; a full encoder retrain without SAP and a larger non-SAP cohort remain stronger tests. Training-time per-language dysarthric-speaker counts per aetiology were not preserved from the April~2026 RunPod runs; Table~I gives per-model language and dysarthric-speaker totals, with per-corpus loading in [2]~\S{}Data. (iii) Pseudo-label circularity: HuBERT is used for both phonological feature extraction and embedding training; pseudo-labelling uses frozen lower layers while embedding training fine-tunes upper layers, partly mitigating the concern. (iv) UA-Speech's shipped manifest is not speaker-disjoint; our LOSO protocol overrides this but the manifest itself may bias evaluations that do not.

\section{Conclusion}
We have shown that per-aetiology contrastive severity embeddings, trained on phonologically pseudo-labelled multilingual data, substantially outperform a mixed-aetiology baseline of identical architecture and training recipe. CP, PD and ALS beat the mixed model by +22.6 \%, +40.0 \% and +32.3 \% relative macro F1 respectively, on filtered, speaker-disjoint, leakage-controlled test pools drawn from each model's own train/val/test partition. To our knowledge, this is among the first controlled cross-lingual comparisons of aetiology-specific severity classifiers for PD and ALS under a matched embedding architecture. Pseudo-labels from a training-free phonological method add value within an aetiology, providing evidence that severity scales are not directly aligned across aetiologies and that pooling them in this training setup is harmful. The practical consequence for the field is that scaling dysarthric-severity data by aetiology-agnostic pooling is counterproductive in this setting; data-pooling strategies should respect the aetiology boundary.

\textbf{Code and data availability.} Code, training manifests and trained model checkpoints will be released to a public repository upon acceptance. The phonological feature pipeline used for pseudo-labelling is the open-source implementation of [1], [2].

\begin{center}
\textbf{Declaration of Generative AI Use}
\end{center}
The first author communicates exclusively via AI-assisted eye-gaze interface due to motor neurone disease. Generative AI (Claude, Anthropic) was used as an assistive communication tool throughout the research process: experimental design, code development, data analysis, statistical computation, figure generation, and manuscript drafting. All scientific decisions, interpretations, and conclusions were made by the authors. The AI served as an accessibility tool enabling a researcher with severe physical disability to conduct computational research --- analogous to a screen reader for visually impaired researchers. No AI-generated content was presented without author review and verification.

\begin{center}
\textbf{Acknowledgments}
\end{center}
CDSD \copyright{} Prof.~Su-Jing Wang, used with permission; SAP under the SAP Terms of Data Use (University of Illinois); portions of the PD data via the OneVoice-MSD~2026 initiative --- we acknowledge the SLT-2026 OneVoice-MSD Focused Track; PC-GITA under a signed academic licence with the GITA laboratory at Universidad de Antioquia (granted to Prof.~A.~A.~Ortiz Barra\~n\'on); HUNDYSDB under an academic licence from the Laboratory of Speech Acoustics, BME; VOC-ALS under the research-only conditions of its Synapse deposit.

\begin{center}
\textbf{References}
\end{center}
\begingroup\sloppy\hbadness=10000
[1] B. Muller, A. A. Ortiz Barra\~n\'on, and L. Roberts, "Training-free cross-lingual dysarthria severity assessment via phonological subspace analysis in self-supervised speech representations," arXiv:2604.10123, 2026. doi: 10.48550/arXiv.2604.10123

[2] B. Muller, A. A. Ortiz Barra\~n\'on, and L. Roberts, "Phonological subspace collapse is aetiology-specific and cross-lingually stable: Evidence from 3,374 speakers," arXiv:2604.21706, 2026. doi: 10.48550/arXiv.2604.21706

[3] P. Enderby, "Disorders of communication: dysarthria," Handbook of Clinical Neurology, vol. 110, pp. 273--281, 2013. doi: 10.1016/B978-0-444-52901-5.00022-8

[4] S. Venugopalan et al., "Speech intelligibility classifiers from 550k disordered speech samples," in Proc. ICASSP, 2023.

[5] K. L. Stipancic et al., "`You say severe, I say mild': Toward an empirical classification of dysarthria severity," J. Speech Lang. Hear. Res., vol. 64, no. 12, pp. 4718--4735, 2021.

[6] B. Kadirvelu et al., "Speaker-independent dysarthria severity classification using self-supervised transformers and multi-task learning," PLOS Digit. Health, vol. 4, no. 11, p. e0001076, 2025.

[7] E. J. Yeo, K. Choi, S. Kim, and M. Chung, "Cross-lingual dysarthria severity classification for English, Korean, and Tamil," in Proc. APSIPA ASC, 2022, pp. 566--574. doi: 10.23919/APSIPAASC55919.2022.9980124

[8] J. Bae et al., "Something from nothing: Data augmentation for robust severity level estimation of dysarthric speech," arXiv:2603.15988, 2026.

[9] W.-N. Hsu et al., "HuBERT: Self-supervised speech representation learning by masked prediction of hidden units," IEEE/ACM Trans. Audio Speech Lang. Process., vol. 29, pp. 3451--3460, 2021.

[10] A. Baevski et al., "wav2vec 2.0: A framework for self-supervised learning of speech representations," in Proc. NeurIPS, 2020, pp. 12449--12460.

[11] A. van den Oord, Y. Li, and O. Vinyals, "Representation learning with contrastive predictive coding," arXiv:1807.03748, 2018. doi: 10.48550/arXiv.1807.03748

[12] F. Schroff, D. Kalenichenko, and J. Philbin, "FaceNet: A unified embedding for face recognition and clustering," in Proc. CVPR, 2015, pp. 815--823. doi: 10.1109/CVPR.2015.7298682

[13] M. McAuliffe et al., "Montreal Forced Aligner: Trainable text-speech alignment using Kaldi," in Proc. Interspeech, 2017, pp. 498--502.

[14] J. R. Duffy, Motor Speech Disorders: Substrates, Differential Diagnosis, and Management, 4th ed. St. Louis: Elsevier, 2019.

[15] E. J. Yeo et al., "Multilingual dysarthric speech assessment using universal phone recognition and language-specific phonemic contrast modeling," in Proc. Interspeech, 2025.

[16] M. Hasegawa-Johnson et al., "Community-supported shared infrastructure in support of speech accessibility," J. Speech Lang. Hear. Res., vol. 67, no. 11, pp. 4162--4175, 2024.

[17] F. Rudzicz, A. K. Namasivayam, and T. Wolff, "The TORGO database of acoustic and articulatory speech from speakers with dysarthria," Language Resources and Evaluation, vol. 46, no. 4, pp. 523--541, 2012. doi: 10.1007/s10579-011-9145-0

[18] H. Kim et al., "Dysarthric speech database for universal access research," in Proc. Interspeech, 2008, pp. 1741--1744.

[19] M. Rusko et al., "Slovak database of speech affected by neurodegenerative diseases," Sci. Data, vol. 11, p. 1320, 2024.

[20] Y. Wan et al., "CDSD: Chinese dysarthria speech database," in Proc. Interspeech, 2024, pp. 4109--4113.

[21] Linguistic Data Consortium, "The SSNCE Database of Tamil Dysarthric Speech," LDC2021S04, 2021. doi: 10.35111/hkh2-vh40

[22] R. Turrisi et al., "EasyCall corpus: A dysarthric speech dataset," in Proc. Interspeech, 2021, pp. 41--45.

[23] J. Mendes-Laureano et al., "NeuroVoz: A Castilian Spanish corpus of parkinsonian speech," Sci. Data, vol. 11, p. 1367, 2024.

[24] L. M. T. Jesus, I. Belo, J. Machado, and A. Hall, "The advanced voice function assessment databases (AVFAD): Tools for voice clinicians and speech research," in Advances in Speech-language Pathology. IntechOpen, 2017. doi: 10.5772/intechopen.69643

[25] R. Dubbioso et al., "Voice signals database of ALS patients with different dysarthria severity and healthy controls," Sci. Data, vol. 11, no. 1, p. 800, 2024.

[26] M. Gao et al., "Enhancing voice wake-up for dysarthria: Mandarin Dysarthria Speech Corpus release and customized system design," in Proc. Interspeech, 2024.

[27] Facebook AI, "facebook/hubert-base-ls960," HuggingFace Model Hub, 2021. \url{https://huggingface.co/facebook/hubert-base-ls960}

[28] F. Pedregosa et al., "Scikit-learn: Machine learning in Python," JMLR, vol. 12, pp. 2825--2830, 2011.

[29] A. Hernandez et al., "Adapting self-supervised speech representations for cross-lingual dysarthria detection in Parkinson's disease," arXiv:2603.22225, 2026.

[30] J. R. Orozco-Arroyave et al., "New Spanish speech corpus database for the analysis of people suffering from Parkinson's disease," in Proc. LREC, 2014, pp. 342--347.

[31] G. Jenei et al., "HUNDYSDB: A Hungarian dysarthric speech database for clinical and technological research," in Proc. TSD, Springer LNCS, 2021.

\endgroup

\end{document}